\documentclass[12pt,onecolumn]{IEEEtran}

\usepackage{cite}
\usepackage{amsmath,amssymb,amsfonts}
\usepackage{algorithmic}
\usepackage{graphicx}
\usepackage{textcomp}
\usepackage{booktabs}
\usepackage{multirow}
\usepackage{array}
\usepackage{makecell}
\usepackage{lettrine}
\usepackage{etoolbox}
\usepackage{xcolor}
\usepackage[hidelinks]{hyperref}

\def\BibTeX{{\rm B\kern-.05em{\sc i\kern-.025em b}\kern-.08em
		T\kern-.1667em\lower.7ex\hbox{E}\kern-.125emX}}

\title{ASM-UNet: Adaptive Scan Mamba Integrating Group Commonalities and Individual Variations for Fine-Grained Segmentation}

\author{
	Bo Wang,
	Mengyuan Xu,
	Yue Yan,
	Yuqun Yang*, \IEEEmembership{Member, IEEE},
	Kechen Shu,
	Wei Ping,
	Xu Tang, \IEEEmembership{Senior Member,~IEEE},
	Wei Jiang,
	Zheng You
	\thanks{Corresponding author: Yuqun Yang (email: yqunyang@163.com).}
}

\begin{document}
	\maketitle

\begin{abstract}
Precise lesion resection depends on accurately identifying fine-grained anatomical structures. While many coarse-grained segmentation (CGS) methods have been successful in large-scale segmentation (e.g., organs), they fall short in clinical scenarios requiring fine-grained segmentation (FGS), which remains challenging due to frequent individual variations in small-scale anatomical structures. Although recent Mamba-based models have advanced medical image segmentation, they often rely on fixed manually-defined scanning orders, which limit their adaptability to individual variations in FGS. To address this, we propose ASM-UNet, a novel Mamba-based architecture for FGS. It introduces adaptive scan scores to dynamically guide the scanning order, generated by combining group-level commonalities and individual-level variations. Experiments on two public datasets (ACDC and Synapse) and a newly proposed challenging biliary tract FGS dataset, namely BTMS, demonstrate that ASM-UNet achieves superior performance in both CGS and FGS tasks. Our code and dataset are available at \href{https://github.com/YqunYang/ASM-UNet}{https://github.com/YqunYang/ASM-UNet}.
\end{abstract}


\section{Introduction}
\label{sec:introduction}
\lettrine{\textbf{R}}ecently, precise resection of the lesion has garnered increasing attention, with the goal of completely removing the lesion while preserving as much healthy tissue as possible. This approach promotes faster and more effective postoperative recovery and reduces the risk of long-term complications~\cite{jiao2024prostate, li2025nir}. To enable precise resection, artificial intelligence technology is employed to automatically segment medical images and construct 3D models, assisting clinicians in accurately identifying the optimal starting point for resection and predicting the extent of lesion and surrounding vessels~\cite{azad2024medical, shaker2024unetr++, yang2025fsvs}. This ultimately facilitates greater surgical precision in the resection of the lesion.

\begin{figure}[!h]
	\centering
	\includegraphics[width=0.65\linewidth]{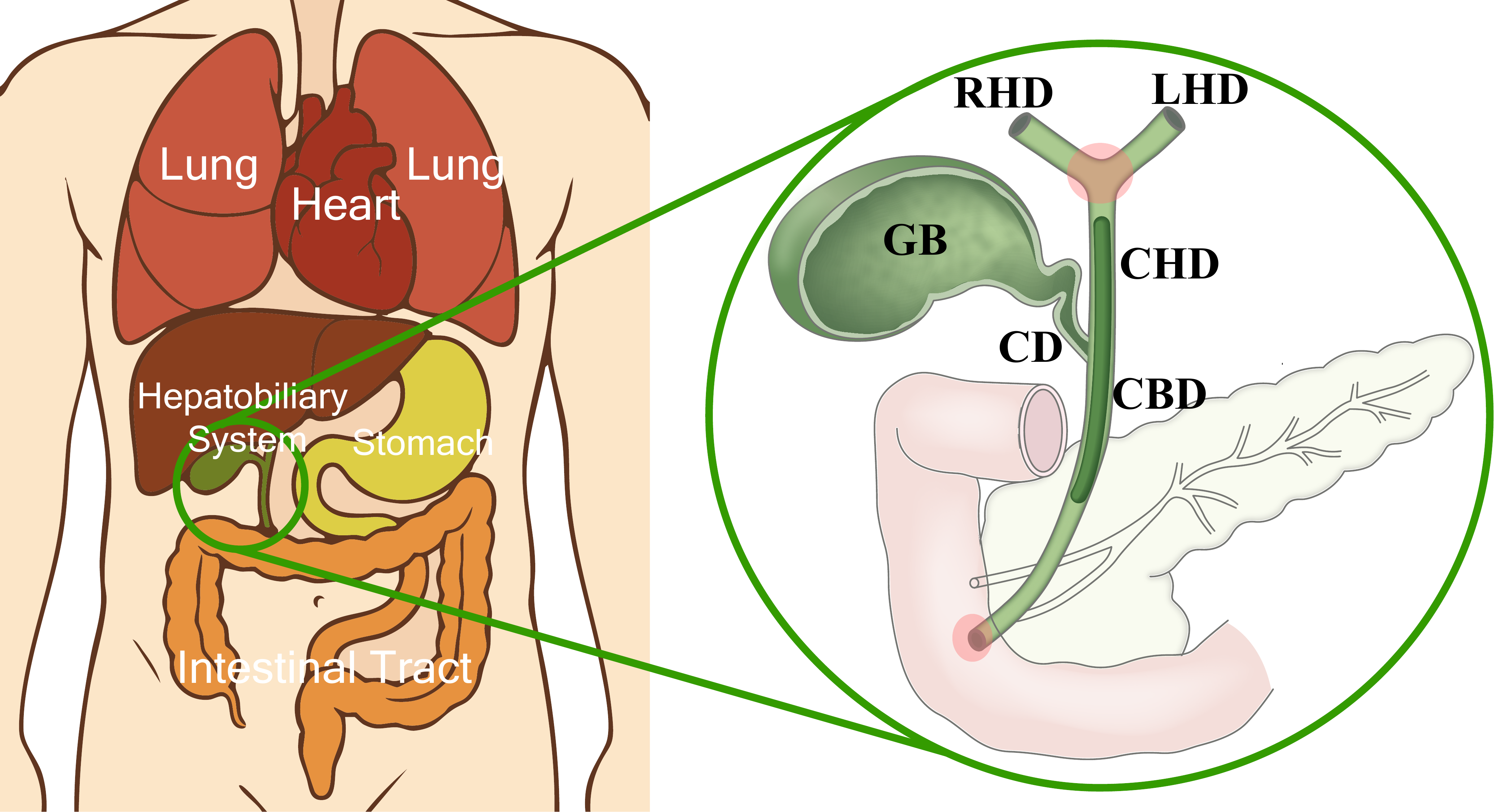}
	\caption{Illustration of fine-grained anatomical structures within the biliary tract system under the coarse-grained organ segmentation. It consists of multiple anatomical structures, including gallbladder (GB), cystic duct (CD) and others. The definitions of all abbreviations are provided in Table~\ref{tbl:class_name}.
		\label{fig:fined_segmentation}}
\end{figure}
In CT image segmentation of organs such as the heart, liver, and lungs, the large size of the imaging area contributes to high segmentation accuracy in terms of total volume \cite{fang2020multi, ma2021abdomenct, liu2024cosst}.  This supports surgical planning for resection procedures involving larger anatomical regions. However, such coarse-grained segmentation (CGS) is insufficient for precise resection. For example, as shown in Fig. \ref{fig:fined_segmentation}, the biliary tract system can be further refined into multiple anatomical structures. During gallbladder resection, it is crucial to accurately distinguish the common hepatic duct (CHD), right hepatic duct (RHD), and common bile duct (CBD) at a fine-grained level to prevent surgical errors. Therefore, achieving accurate fine-grained segmentation (FGS) of medical images has become a critical topic in precision surgery. 

\begin{figure}[t]
	\centering
	\includegraphics[width=1\linewidth]{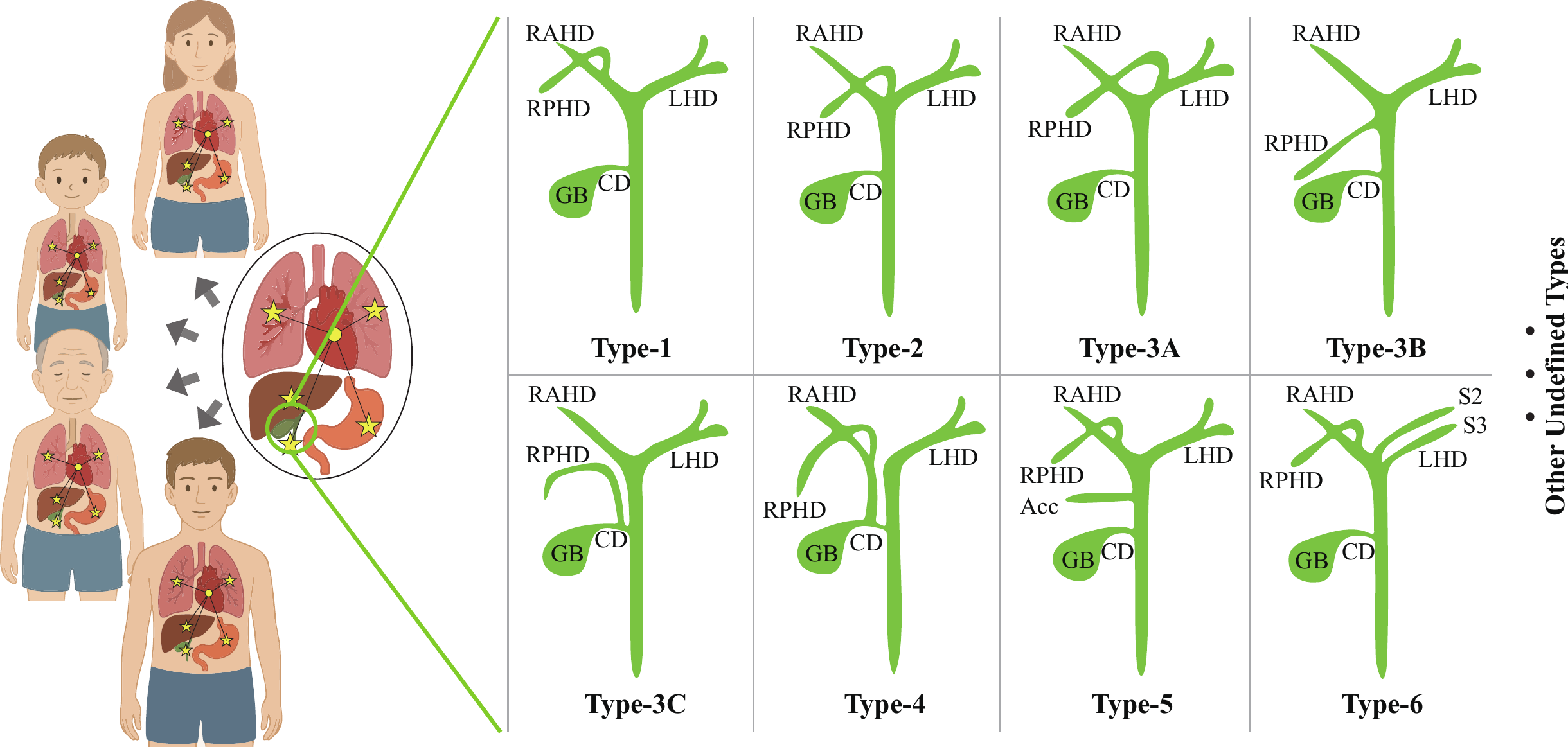}
	\caption{Illustration of six common anatomical variants of the biliary tract system. Beyond these, numerous undefined variants also occur in clinical practice.
		\label{fig:duct_classification}}
\end{figure}

Nevertheless, individual anatomical variation remains a major challenge in FGS. As illustrated in the left part of Fig. \ref{fig:duct_classification}, group-level commonality, such as the general distribution patterns of most organs, is consistently observed across individuals of different ages and sexes, which facilitates large-scale CGS. In contrast, when the anatomical scale is reduced from the entire thoracoabdominal cavity to single organs or functional systems, inter-individual differences become more pronounced. Taking the biliary system as an example, it can be classified into six common types \cite{nakamura2002anatomical, varotti2004anatomic}, as shown in the right part of Fig. \ref{fig:duct_classification}, along with other undefined variants. These variations correspond to different distributions of anatomical structures such as RHD and CHD, which are critical components for precise resection. Moreover, such variations often occur at a small anatomical scale. For example, the CHD typically has a diameter of less than 10 mm.  Therefore, although achieving accurate FGS is essential for precision surgery, it remains a challenging task due to significant individual anatomical variations.

Among the various approaches proposed for medical image segmentation, convolutional neural network (CNN)-based models have become the dominant paradigm~\cite{kayalibay2017cnn}. In particular, U-Net~\cite{ronneberger2015u} and its variants~\cite{zhou2018unet++, huang2020unet, wang2022uctransnet} have achieved encouraging performance, benefiting from their encoder–decoder architecture and multi-scale feature fusion. However, CNNs inherently struggle to capture long-range dependencies, limiting their effectiveness in segmenting complex anatomical structures. To address this issue, researchers have introduced the Transformer architecture~\cite{vaswani2017attention}. With its self-attention mechanism, the Transformer excels at modeling global context and long-range dependencies, making it particularly well-suited for tasks that require a holistic understanding of medical images~\cite{oktay2018attention}. As a result, many Transformer-involved methods~\cite{huang2022missformer, zhou2023nnformer, wang2025biclip} have been proposed, achieving promising accuracy across a variety of medical image segmentation tasks. Recently, to address the high computational complexity of Transformers, which scales quadratically with input length ($\mathcal{O}(n^2)$), Mamba~\cite{gu2023mamba} has been introduced. Mamba leverages a selective state space model (SSM) and linear-time ($\mathcal{O}(n)$) operations, enabling efficient parallel processing while preserving the capability to model long-range dependencies. Building on these advantages, researchers have adopted the Mamba architecture to develop medical image segmentation models~\cite{gong2025nnmamba, xing2024segmamba, liu2024swin}, achieving both effective and efficient performance.

However, challenges remain in applying these models to medical imaging tasks. Unlike text, which follows a one-dimensional sequential order (e.g., forward or backward), medical image data is inherently three-dimensional and can be scanned in multiple directions, such as horizontal, vertical, or diagonal~\cite{zhu2024vision, huang2024localmamba, ren2024mambacsr}. Although scanning-based feature extraction can effectively emphasize anatomical details, existing methods typically rely on manually designed and fixed scanning orders. This rigidity limits their adaptability to the diverse and complex characteristics of FGS, which often exhibit intricate structures, irregular shapes, and varying spatial positions. For example, accurately identifying CHD often requires scanning along RHD and LHD to locate their intersection, where an appropriate scanning order helps ensure precise localization. However, due to individual variations, the fixed scanning orders used in existing Mamba-based methods struggle to follow this identification process effectively. Therefore, how to generate adaptive scanning orders for the different individuals has become an important problem.

To address this issue, we propose a novel adaptive scanning mechanism that predicts a scan score for each medical image (i.e., each individual), which is then embedded into the feature representation. The scan order is dynamically determined by sorting these scores, allowing the model to follow adaptive scanning paths. Considering the group-level commonality in large anatomical scale and individual-level variation in smaller ones, we design a dual-scan scoring strategy: a group scan score shared across all images and an individual scan score generated specifically for each image. These two scores are integrated to jointly and adaptively determine the final scan order, enabling the model to effectively capture both group-level commonality and individual-level variation, thereby improving segmentation accuracy at both coarse- and fine-grained levels. To evaluate the performance of our method, we conduct experiments on two public datasets, Synapse multi-organ segmentation (Synapse)\cite{landman2015miccai} and the automatic cardiac diagnosis challenge (ACDC)\cite{bernard2018deep}, as well as on our proposed biliary tract multi-class segmentation (BTMS) dataset for FGS. The BTMS dataset poses a significant segmentation challenge due to its eight fine-grained categories within the small and complex anatomical structures of the biliary tract system (see Figure~\ref{fig:fined_segmentation}).
The main contributions of this work are as follows:
\begin{itemize}
	\item We propose an  adaptive scanning Mamba (ASM) that dynamically determines scan paths by jointly considering group-level commonality and individual-level variation, enhancing segmentation performance.
	\item We introduce a biliary tract multi-class segmentation dataset, BTMS, which includes eight categories at small anatomical scales, serving as a meaningful and challenging benchmark for FGS.
	\item We conduct extensive experiments on two public datasets and our BTMS dataset, demonstrating the effectiveness and generalizability of the proposed method on both coarse- and fine-grained segmentation tasks.
\end{itemize}

\section{Related Work}
\label{sec:related_works}
To provide a comprehensive overview, this section covers two key areas relevant to our study: common approaches in medical image segmentation and recent advances in Mamba-based methods.

\subsection{Medical Image Segmentation}
Fully Convolutional Networks (FCNs)~\cite{long2015fully} use convolution and upsampling for pixel-level prediction but lack sufficient skip connections, making it hard to retain spatial details. To address this issue, U-Net~\cite{ronneberger2015u} adopts a symmetric encoder-decoder architecture with skip connections at each scale, effectively combining low-level local information with high-level semantic features. It has since become a strong backbone in medical image segmentation. Since then, many variants of U-Net have been proposed. For example, U-Net++~\cite{zhou2018unet++} refines skip connections to enhance multi-scale feature aggregation, while U-Net3+~\cite{huang2020unet} introduces full-scale skip connections to improve feature integration. Among them, nnU-Net~\cite{isensee2021nnu} stands out as a self-configuring framework that automatically adapts network architecture, training, and preprocessing to a given dataset, achieving strong performance without manual tuning. To fully explore the potential of CNNs, ConvNeXt~\cite{liu2022convnet} modernizes the ResNet architecture with a series of design improvements, achieving strong performance in image segmentation while retaining the simplicity and efficiency of convolutional models. Despite these advancements, modeling long-range dependencies remains difficult, prompting the introduction of Transformers to boost performance. For example, MissFormer~\cite{huang2022missformer} leverages a Transformer architecture to effectively integrate local context and global dependencies, enhancing performance in medical image segmentation. Similarly, CoTr~\cite{xie2021cotr} and nnFormer~\cite{zhou2023nnformer} integrate CNNs with Transformers to achieve accurate medical image segmentation. In UNETR~\cite{hatamizadeh2022unetr}, the Transformer is employed as the encoder to capture global multi-scale contextual information. Building on this framework, several variants such as Swin UNETR~\cite{hatamizadeh2021swin} and UNETR++~\cite{shaker2024unetr++} have been proposed, achieving promising segmentation performance. While Transformers have demonstrated strong capability in modeling global dependencies, their inherent computational complexity leaves considerable room for more efficient solutions.

\subsection{Mamba-based Methods}

Mamba~\cite{gu2023mamba} was proposed to address key limitations of Transformer models, such as quadratic computational complexity and inefficiency in modeling long-range dependencies. By leveraging SSMs, Mamba achieves linear complexity while maintaining strong performance through efficient sequence modeling. As a result, various Mamba-based methods have been developed for medical image segmentation. For instance, VM-UNet~\cite{ruan2024vm} incorporates a visual state space block into the U-shaped architecture to effectively capture global contextual information. Building upon Mamba, Swin-UMamba~\cite{liu2024swin1} exploits the pretraining capability of ImageNet to enhance segmentation performance. Its variant, Swin-UMamba~\cite{liu2024swin}, further adopts a self-supervised learning strategy to bridge the domain gap between natural and medical images, thereby significantly improving generalization on small datasets while maintaining strong performance across multiple benchmarks. These methods have demonstrated remarkable effectiveness in 2D medical image segmentation. Furthermore, Vivim~\cite{yang2024vivim} applies Mamba to video segmentation with a temporal module to compress long sequences. In addition to the 2D images, researchers extend to employing Mamba for segmenting 3D medical images. For instance, U-Mamba~\cite{ma2024u} combines the local feature extraction capabilities of CNNs with the long-range dependency modeling strength of SSMs. Following this trend, NnMamba~\cite{gong2024nnmamba} further unifies CNNs and SSMs to effectively capture long-range dependencies in 3D medical image analysis. SegMamba~\cite{xing2024segmamba} introduces the tri-directional spatial Mamba (TSMamba) module to capture multi-scale global information, while LKM-UNet~\cite{wang2024lkm} enhances local modeling through large kernel Mamba blocks, expanding the receptive field for more precise spatial feature extraction. Although these Mamba-based methods demonstrate promising performance in medical image segmentation, their reliance on a fixed scanning order still limits generalization in fine-grained segmentation tasks, which is the problem our method is designed to solve.

\section{Dataset}
\label{sec:dataset}
The biliary system plays a critical role in the human digestive system, with primary functions including the transport and storage of bile produced by the liver, the regulated release of bile into the duodenum, and the prevention of intestinal content reflux. It comprises several important anatomical structures, including the CB, CD, CBD, CHD, RHD, RPHD, and RAHD, whose abbreviations are explained in Table ~\ref{tbl:class_name}.
\begin{table}[!h]
	\caption{Anatomical Categories with Abbreviations. Categories Marked with ``*'' Are Challenging to Segment}
	\centering
	\label{tbl:class_name} 
	\renewcommand{\arraystretch}{1.2}
	\setlength{\tabcolsep}{4mm}
	{\begin{tabular}{c|c|c}
			\toprule
			ID & Category Name & Abbreviation\\
			\midrule
			0 & Background & - \\
			1 & Gallbladder & CB\\
			2 & $\text{Cystic Duct}^{\ast}$ & $\text{CD}^{\ast}$\\
			3 & Common Bile Duct & CBD\\
			4 & Common Hepatic Duct & CHD\\
			5 & $\text{Right Hepatic Duct}^{\ast}$ & $\text{RHD}^{\ast}$\\
			6 & Right Posterior Hepatic Duct & RPHD\\
			7 & Right Anterior Hepatic Duct & RAHD\\
			8 & Left Hepatic Duct & LHD\\
			\bottomrule
	\end{tabular}}
\end{table}

\begin{table*}[t]
	\centering
	\caption{Statistics of Publicly Available Datasets}
	\label{tbl:statistical_info}
	% \renewcommand{\arraystretch}{1.2} % 调整行间距
	% \footnotesize % 使用较小的字体
	% \setlength{\tabcolsep}{3mm} % 调整列间距
	\resizebox{\textwidth}{!}{
		\begin{tabular}{c|c|cc|cc|cc|c}
			\toprule
			\multirow{2}{*}{Dataset}  &   \multirow{2}{*}{Resolution}  &\multicolumn{2}{c|}{Number} & \multicolumn{2}{c|}{Slice thickness} & \multicolumn{2}{c|}{Slice number} & \multirow{2}{*}{Annotated Biliary Classes}\\
			\cline{3-8}
			\noalign{\vskip 0.6ex}
			& & Cases & Images & Range & Medium  & Range & Medium  & \\
			\midrule
			3Dircadb1~\cite{soler20103d} & 512$\times$512 & 20 & 3,209 & [1.0, 4.0] & 1.6 & [74, 260] & 130 & GB \\
			Synapse~\cite{landman2015miccai} & 512$\times$512 & 50 & 6,811 & [2.5, 5.0] & 3.0 & [85, 198] & 127 & GB \\
			WORD~\cite{luo2022word} & 512$\times$512 & 150 & \textbf{30,495} & [2.5, 3.0] & 3.0 & [151, 343] & 202 & GB\\
			%			3D GRASE MRCP*~\cite{oh20243d} & &\textbf{200} & & & & & & {GB, CD, CBD, CHD, IHD}\\
			\midrule
			\multirow{2}{*}{BTMS} & \multirow{2}{*}{512$\times$512} & \multirow{2}{*}{100} & \multirow{2}{*}{20,419} & \multirow{2}{*}{[\textbf{0.625}, \textbf{1.5}]} & \multirow{2}{*}{\textbf{1.2}} & \multirow{2}{*}{[\textbf{169}, \textbf{729}]} & \multirow{2}{*}{\textbf{269}} & \textbf{GB, CD, CBD, CHD, } \\ & & & & & & & & \textbf{RHD, RPHD, RAHD, LHD} \\
			\bottomrule
		\end{tabular}
	}
\end{table*}

These anatomical structures exhibit substantial complexity and variability due to individual anatomical variations. Precise delineation of intrahepatic bile ducts (e.g., RPHD, RAHD, LHD), extrahepatic bile ducts (e.g., CHD, CBD), and associated structures (e.g., CB) is essential for various clinical applications, including preoperative planning in liver transplantation, minimizing surgical complications during laparoscopic hepatectomy, and improving diagnostic accuracy in biliary diseases. Therefore, fine-grained segmentation of the biliary system holds great clinical significance.

\begin{figure}[t]
	\centering
	\includegraphics[width=0.6\linewidth]{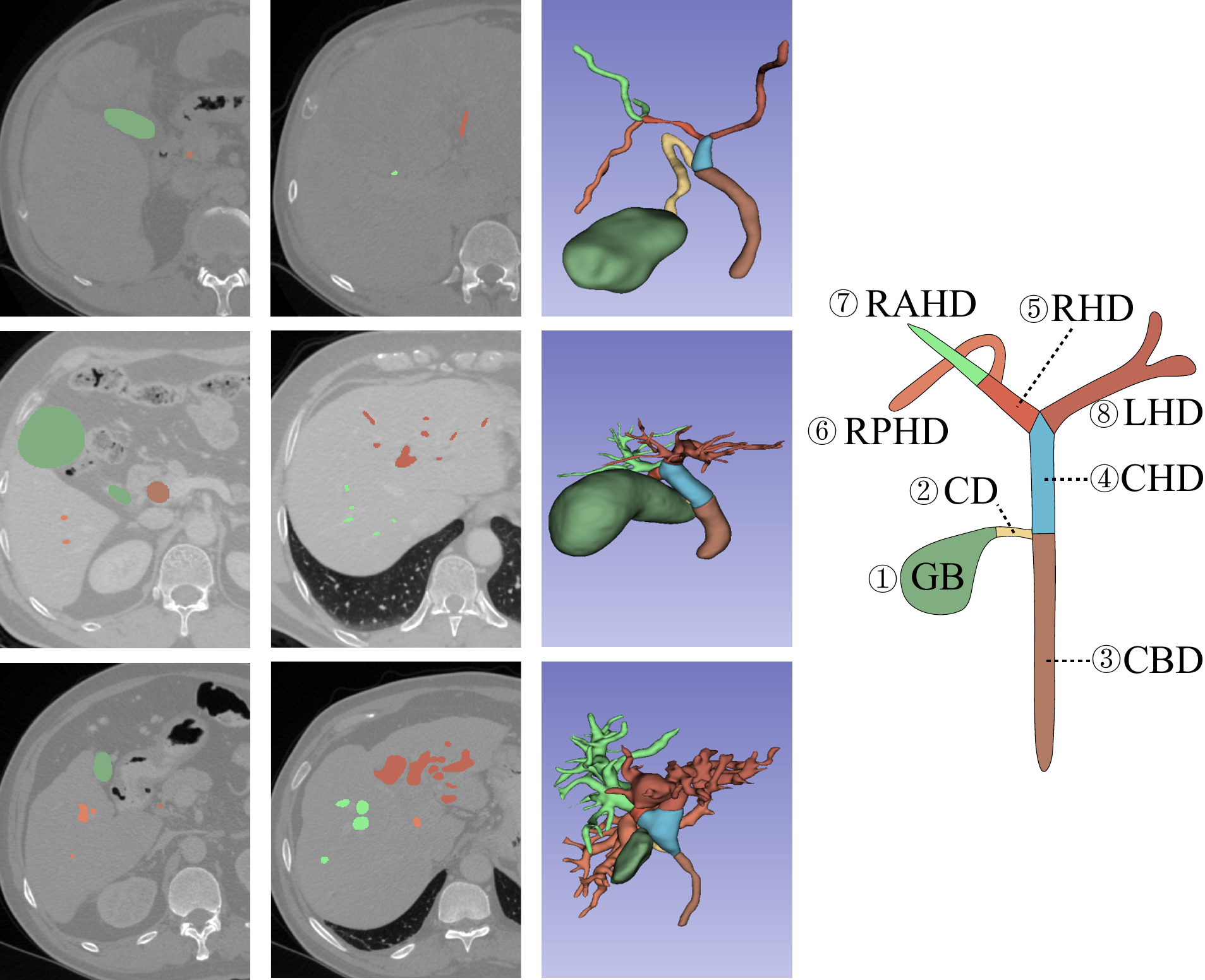}
	\caption{Examples from our BTMS dataset. Each row corresponds to a single case, with the right section providing a simplified explanation of color coding and anatomical location.}
	\label{fig:dataset}
\end{figure}

However, existing publicly available datasets for the biliary system remain at a coarse-grained level. As shown in Table \ref{tbl:statistical_info}, all the datasets (e.g., 3Dircadb1~\cite{soler20103d}, Synapse~\cite{landman2015miccai}, and WORD~\cite{luo2022word} ) only provide annotations for the gallbladder (GB), lacking detailed segmentation of the complete biliary tree. In contrast, our BTMS dataset offers fine-grained annotations for eight biliary structures, enabling more comprehensive anatomical understanding. Moreover, BTMS contains 100 cases with 20,419 images, which is significantly larger than 3Dircadb1 (20 cases / 3,209 images) and Synapse (50 cases / 6,811 images), and comparable in scale to WORD (150 cases / 30,495 images) but with much finer anatomical labeling. Additionally, BTMS has higher-resolution slices (median thickness: 1.2 mm) and more slices per case (median: 269), providing richer spatial detail for accurate biliary system analysis. Three case examples, along with the explanation of color coding and anatomical locations, are shown in Fig. \ref{fig:dataset}. In the experiments, the dataset is split into training, validation, and test sets in a 4:3:3 ratio. 

Notably, CD and RHD are marked with “*” in Table~\ref{tbl:class_name} due to their significant segmentation difficulty, which arises from anatomical variations and their small-scale structures. For example, CD is extremely short and thin, making it difficult to locate. RHD shares similar characteristics, and its segmentation is even more challenging because its position can vary considerably or even disappear (e.g., Type-2 in Figure~\ref{fig:duct_classification}) entirely due to changes at the intersection of RAHD and RPHD, further increasing the difficulty compared to CD.

\begin{figure*}[t]
	\centering
	\includegraphics[width=0.85\textwidth]{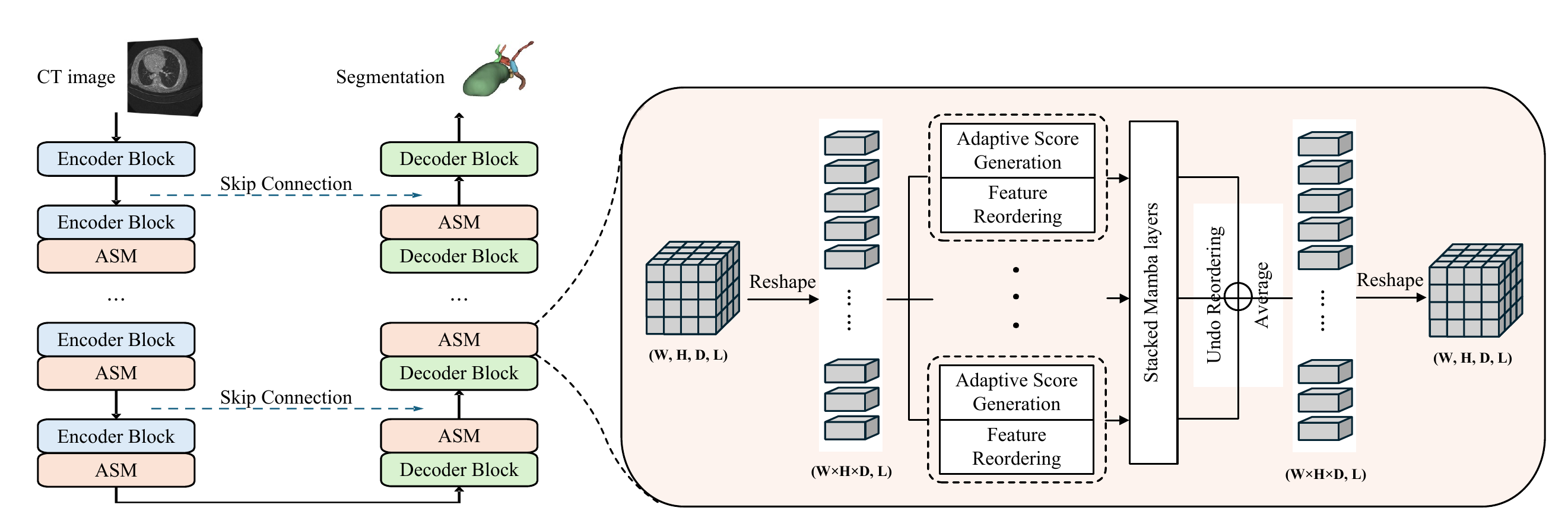}
	\caption{Overview of the proposed network architecture ASM-UNet.
    \label{fig:overview}}
\end{figure*}

\begin{figure*}[!h]
	\centering
	\includegraphics[width=0.85\textwidth]{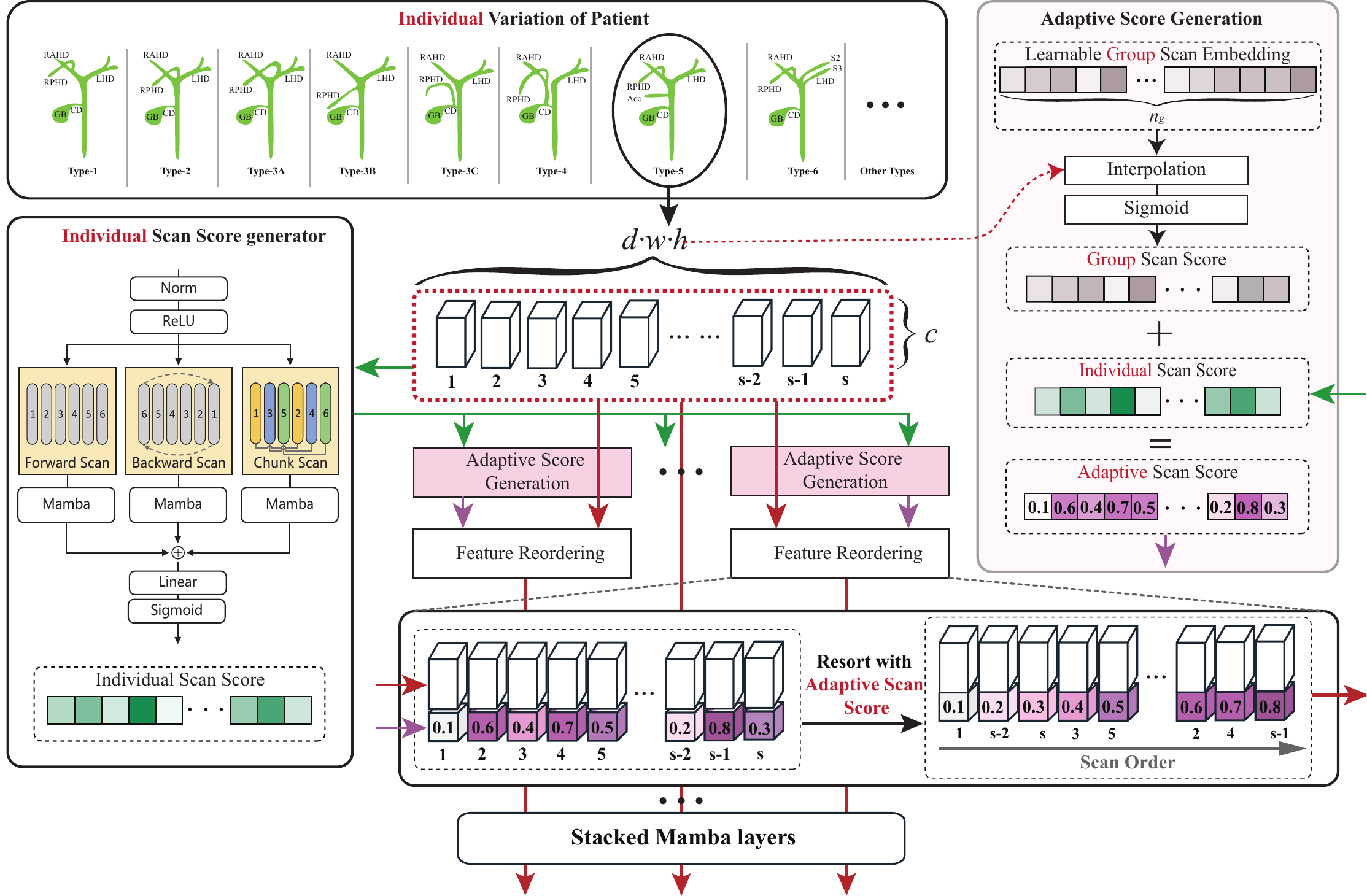}
	\caption{Diagram of adaptive score generation and feature reordering. The adaptive scan score is obtained by combining both the group and individual scan scores, and is used to adaptively guide the scanning order.
		\label{fig:asm}}
\end{figure*}

\section{Methodology}
\label{sec:methodology}

\subsection{Overview}
\label{sub:unet}
As shown in Figure~\ref{fig:overview}, the proposed model is built upon the U-Net architecture, following the design principles of the literature \cite{isensee2021nnu}, and consisting of six encoder blocks, six decoder blocks, and adaptive scanning Mamba (ASM) blocks. Note that the ASM block is inserted after the first encoder block and before the last decoder block. This design choice is motivated by the fact that feature maps at these stages have large spatial dimensions (corresponding to very long sequences) and limited channel information, making it difficult to capture global context effectively while incurring high computational cost.
A three-dimensional medical image with dimensions $W \times H \times D$ is used as the input to the model, while the output is a segmentation map of shape $W \times H \times D \times C$. Here, $W$, $H$, $D$, and $C$ represent the width, height, depth, and number of segmentation categories, respectively.
During data flow, the downsampling rates and output channels of the six encoder blocks are set to $[1, 2, 2, 2, 2]$ and $[32, 64, 128, 256, 320]$, respectively, and those of the decoder blocks are configured correspondingly. Specifically, each encoder block consists of two convolutional layers, followed by normalization and ReLU activation, with a kernel size of $3 \times 3 \times 3$. The first convolutional layer uses a stride of 2 for downsampling, while the second uses a stride of 1.
Decoder blocks perform upsampling using transposed convolution, followed by concatenation with the corresponding feature map from the encoder block via skip connection. The combined feature map is then processed through two convolutional layers (with $3 \times 3 \times 3$ kernels and stride 1), followed by normalization and ReLU activation, to produce the input for the next decoder block.

\subsection{Adaptive Scanning Mamba}
\label{sub:asm}

The ASM module is designed to enhance generalization performance by employing an adaptive scanning mechanism instead of a fixed one. As illustrated in Fig.~\ref{fig:overview}, the process begins by flattening the feature map along the spatial dimensions (width, height and depth), i.e., from $(W, H, D, L)$ to $(W\times H\times D, L)$, which is then fed into several blocks containing adaptive score generation and feature reordering modules. Then, the reordered feature maps are passed through a series of stacked Mamba layers. Next, the multiple output feature maps are restored to their original order (i.e., undo reordering) and averaged at the pixel level. Finally, the averaged feature map is reshaped along the spatial dimensions to produce the output of the ASM module. Here, the adaptive scan score for guiding the scan order is generated by adding group scan score and individual scan score. 

\subsubsection{Group scan score}
Since group-level commonalities are consistently observed among individuals, such as the spatial distribution and volume of most organs, emphasizing these patterns can help the model better capture the global anatomical context. Therefore, a group scan score is defined to guide the scanning order at the global level. Since it aims to leverage group-level commonalities without being influenced by individual-specific characteristics, the score is generated through a learnable parameter within the model, namely the learnable group scan embedding, as illustrated in the adaptive score generation block of Fig.~\ref{fig:asm}. The length $n_g$ of the embedding is a hyperparameter, and is initialized as a vector sampled from a normal distribution. To accommodate feature maps of arbitrary spatial size, the embedding is expanded and adjusted based on the spatial dimensions ($d \cdot w \cdot h = s$) using linear interpolation during feature learning. Finally, the Sigmoid function is applied to normalize the embedding into a score, ensuring that its values lie within the range $[0, 1]$.

\subsubsection{Individual scan score}
Since small-scale anatomical structures often exhibit variability, they contribute to differences between individuals, i.e., individual variation. Therefore, accurately capturing individual-specific information is essential for performing the FGS task. To achieve this, an individual scan score generator is designed, as shown in Fig.~\ref{fig:asm}. It takes as input the feature maps of a specific patient and generates an individual scan score by comprehensively analyzing the medical image using three different fixed scan orders \cite{gu2023mamba} to ensure generation accuracy. The length of the generated score is the same as that of the input feature maps, i.e., $d \cdot w \cdot h = s$. This design is inspired by the observation that, when clinicians examine medical images, an initial fixed scanning order (for example, from top to bottom or from right to left) helps them memorize the anatomical content at each location. Based on this internal memory, they can perform dynamic scanning of regions potentially related to disease, more effectively analyze their relationships, and draw diagnostic conclusions. In our method, three fixed scan orders serve as the model's initial scanning phase, similar to how clinicians first build an understanding of anatomical structures such as the left and right hemi-liver, which are important for identifying the RHD and LHD. Based on this global understanding, the individual scan score is then generated to guide a more adaptive scanning process for accurately identifying small-scale structures such as the CHD and CD, ultimately improving the performance of the FGS task. After both group and individual scan scores are generated, the adaptive scan score can be obtained by adding them. The adaptive scan score is concatenated with the corresponding feature embedding for adaptively guiding the scanning order.

\section{Experiment}
\label{sec:experiment}

\subsection{Datasets}
\label{subsec:datasets}
To comprehensively evaluate the effectiveness and generalizability of our method, namely ASM-UNet, we conduct experiments on two widely-used public datasets ACDC~\cite{bernard2018deep} and Synapse~\cite{landman2015miccai} for medical image segmentation, and our fine-grained BTMS dataset introduced in Section \ref{sec:dataset}. For ACDC dataset, it serves as a benchmark for cardiac segmentation, containing annotations for three anatomical structures: the right ventricle (RV), myocardium (MYO), and left ventricle (LV). It consists of 2,978 slices from 150 cases, in which training and test sets contain 100 and 50 cases, respectively. For Synapse dataset, it provides a diverse collection of annotated medical images for multi-organ segmentation, including eight structures: aorta (AO), gallbladder (GB), left kidney (LK), right kidney (RK), liver (LIV), pancreas (PAN), spleen (SPL) and stomach (STO), comprising a total of 3,779 slices from 30 patients.

\subsection{Data Preprocessing and Implementation Details}
\label{subsec:pre}
All images are resampled to an isotropic resolution based on the median voxel spacing of the dataset. Intensity normalization is performed using z-score standardization, where the mean and standard deviation are computed from the foreground voxels. Data augmentation techniques are applied, including random flipping with a probability of 50\%, random rotation up to 30 degrees, and random scaling with a factor ranging from 0.7 to 1.4. Training is conducted using a patch-based strategy with a batch size of 2. The optimization process employs the Adam optimizer with an initial learning rate of 0.01, which follows a polynomial decay schedule. The loss function consists of a combination of dice loss and cross-entropy loss with equal weighting. The model is trained for a maximum of 1000 epochs, with validation performed every 50 epochs. If no improvement is observed for 60 consecutive validation steps, early stopping is applied to terminate training. To ensure fair comparisons, the hyperparameters of the comparative methods are set as closely as possible to those reported in their original papers. All experiments are implemented using PyTorch and executed on two NVIDIA RTX 3090 GPUs, each equipped with 24 GB of memory.

\begin{table*}[t]
	\caption{Comparison with Alternative Methods on Two Public Datasets}
	\footnotesize
	\centering
	\label{tbl:res_comp} 	
	\renewcommand{\arraystretch}{1.2}
	\setlength{\tabcolsep}{2mm}{
		\begin{tabular}{c|c|ccc||c|cccccccc}
			\toprule
			\multirow{2.5}{*}{\makecell[c]{Methods}}&\multicolumn{4}{c||}{ACDC}  & \multicolumn{9}{c}{Synapse}\\
			\cmidrule(r){2-14}&Avg.&LV&RV&MYO&Avg. &  AO &  GB &  LK   & RK  & LIV  & PAN  & SPL  & STO\\
			\midrule
			VIT+CUP~\cite{chen2024transunet} &81.45	&81.46	&70.71	&92.18	&67.86	&70.19	&45.10	&74.70	&67.40	&91.32	&42.00	&81.75	&70.44\\
			R50-VIT+CUP~\cite{chen2024transunet}&87.57	&86.07	&81.88	&94.75	&71.29	&73.73	&55.13	&75.80	&72.20	&91.51	&45.99	&81.99	&73.95\\
			TransUNet~\cite{chen2024transunet}&89.71	&88.86	&84.54	&\underline{95.73}	&77.48	&87.23	&63.16	&81.87	&77.02	&94.08	&55.86	&85.08	&75.62\\
			Swin-UNet~\cite{cao2022swin}&90.00	&88.55	&85.62	&\textbf{95.83}	&79.13	&85.47	&66.53	&83.28	&79.61	&94.29	&56.58	&90.66	&76.60\\
			% TransClaw UNet~\cite{chang2021transclaw}&&&&&	78.09	&85.87	&61.38	&84.83&	79.36&	94.28	&57.65	&87.74	&73.55\\
			LeVit-UNet-384s~\cite{xu2023levit}&90.32	&89.55	&87.64&	93.76	&78.53	&87.33	&62.23	&84.61	&80.25	&93.11	&59.07	&88.86	&72.76\\
			MISSFormer~\cite{huang2022missformer}&88.27&85.78&85.35&93.68 & 81.96 & 86.99 & 68.65 & 85.21 & 82.00 & 94.41 & 65.67 & 91.92 & 80.81 \\
			CoTr~\cite{xie2021cotr}&90.63&89.72&87.45&94.72 & 80.78 & 85.42 & 68.93 & 85.45 & 83.62 & 93.89 & 63.77 & 88.58 & 76.23 \\
			UNETR~\cite{hatamizadeh2022unetr}&88.61 & 85.29 & 86.52 & 94.02  & 79.56 & 89.99 & 60.56 & 85.66 & 84.80 & 94.46 & 59.25 & 87.81 & 73.99 \\
			Swin UNETR~\cite{hatamizadeh2021swin}&89.41&87.54&86.30&94.38 & 83.51 & 90.75 & 66.72 & 86.51 & 85.88 & 95.33 & 70.07 & \underline{94.59} & 78.20 \\
			nnFormer~\cite{zhou2023nnformer}&\underline{92.06} & \underline{90.94} & \underline{89.58} & 95.65 & \underline{86.57} & \underline{92.04} & \underline{70.17} & \underline{86.57} & 86.25 & \textbf{96.84} & \textbf{83.35} & 90.51 & \textbf{86.83} \\
			SegFormer3D~\cite{perera2024segformer3d}&90.96	&88.50	&88.86	&95.53&82.15&90.43&55.26&86.53&86.13&95.68&73.06&89.02&81.12\\
			\midrule
			\textbf{ASM-Unet}&\textbf{92.61} & \textbf{92.61} & \textbf{90.91} & 94.75 & \textbf{87.27} & \textbf{92.98} & \textbf{73.35} & \textbf{87.15} & \textbf{87.43} & \underline{96.37} & \underline{82.87} & \textbf{95.13} & \underline{82.82} \\
			\bottomrule
		\end{tabular}
	}
	% \vspace{-6mm}
\end{table*}

\begin{table*}[t]
	\caption{Comparison with Alternative Methods on Our BTMS Dataset. $\dag$ Means That the Method Is Mamba-based}
	\footnotesize
	\centering
	\label{tbl:comp_BTMS} 	
		\begin{tabular}{c|c|c|c|cccccccc}
			\toprule
			Methods  & \makecell{Avg. Coarse}  &\makecell{Avg.  Fine \\(w/o Hard)} & \makecell{Avg. Fine\\(w/ Hard)}  &GB	&CD*	&CBD	&CHD	&RHD*	&RPHD	&RAHD	&LHD\\
			\midrule
			VIT-CUP~\cite{chen2024transunet} &39.98 & 17.16 & 12.87 & 45.19 & - & 4.88 & 16.04 & - & 10.24 & 12.62 & 14.01 \\
			R50-VIT-CUP~\cite{chen2024transunet} &48.49  & 21.97 & 20.48 & 51.38 & 5.97 & 5.63 & 16.50 & {26.09} & 16.75 & 21.95 & 19.61 \\
			TransUNet~\cite{chen2024transunet} &60.92& 31.92 & 28.99 & 69.65 & 7.46 & 15.72 & 28.31 & \textbf{32.99} & 22.79 & 26.46 & 28.56 \\
			Swin-UNet~\cite{cao2022swin} &57.11& 25.91 & 19.43 & 60.54 & - & 10.67 & 16.58 & - & 20.73 & 22.79 & 24.15 \\
			CoTr~\cite{xie2021cotr} &74.14 & \underline{46.87} & {40.53} & 80.43 & {23.04} & \underline{47.66} & {48.17} & 19.92 & 30.39 & 32.58 & \underline{42.01} \\
			UNETR~\cite{hatamizadeh2022unetr} & 64.99 & 37.26 & 31.31 & 65.98 & 11.58 & 38.31 & 31.84 & 15.31 & 24.65 & 29.76 & 33.01 \\
			Swin UNETR~\cite{hatamizadeh2021swin} & 62.40 & 36.31 & 30.21 & 69.53 & 8.12 & 33.23 & 27.68 & 15.74 & 26.76 & 28.29 & 32.35 \\
			nnFormer~\cite{zhou2023nnformer} & 67.38 & 41.42 & 36.16 & 68.10 & 19.59 & 40.58 & 40.93 & 21.12 & 29.91 & 31.98 & 37.04 \\
			SegFormer3D~\cite{perera2024segformer3d} & 64.28 & 37.33 & 30.50 & 68.52 & 8.62 & 39.51 & 31.44 & 11.39 & 25.40 & 28.08 & 31.04 \\
			\midrule
			SegMamba$\dag$~\cite{xing2024segmamba} & \underline{74.65} & 42.94 & 35.75 & 79.49 & 11.54 & 40.77 & 42.29 & 16.88 & 27.34 & 30.03 & 37.69 \\
			NnMamba$\dag$~\cite{gong2024nnmamba} & 71.92 & 43.86 & 37.51 & 79.18 & 14.89 & 39.58 & 43.46 & 21.99 & 30.12 & 31.11 & 39.71 \\
			U-Mamba$\dag$~\cite{ma2024u} & 74.20 & 46.23 & \underline{41.69} & \underline{83.37} & \textbf{29.31} & 43.82 & 45.77 & \underline{26.82} & 29.87 & \textbf{33.47} & 41.10 \\
			LKM-UNet$\dag$~\cite{wang2024lkm} & 72.41 &45.40&40.73&77.74&\underline{26.85}&42.65&\underline{49.23}&26.57&\underline{31.51}&29.47&41.79\\
			\midrule
			\textbf{ASM-UNet$\dag$} & \textbf{76.74} & \textbf{48.79} & \textbf{42.57} & \textbf{84.59} & {23.41} & \textbf{48.36} & \textbf{51.60} & 24.37 & \textbf{32.51} & \underline{32.70} & \textbf{42.98} \\
			% \hline
			\bottomrule
		\end{tabular}
		% \vspace{-6mm}
	\end{table*}

\subsection{Comparison Studies}
\label{subsec:compare}

We perform a comprehensive comparison of our method with thirty other models on three datasets. The comparative methods consist of the nine widely-used methods TransUNet~\cite{chen2024transunet}, Swin-UNet~\cite{cao2022swin}, LeViT-UNet~\cite{xu2023levit}, MISSFormer~\cite{huang2022missformer}, CoTr~\cite{xie2021cotr}, UNETR~\cite{hatamizadeh2022unetr}, Swin UNETR~\cite{hatamizadeh2021swin}, nnFormer~\cite{zhou2023nnformer}, and SegFormer3D~\cite{perera2024segformer3d}, as well as four Mamba-based methods SegMamba~\cite{xing2024segmamba}, NnMamba~\cite{gong2024nnmamba}, U-Mamba~\cite{ma2024u}, LKM-UNet~\cite{wang2024lkm}. The quantitative results based on the Dice metric are reported in Table~\ref{tbl:res_comp} and Table~\ref{tbl:comp_BTMS}.

\textit{ACDC and Synapse}: Table~\ref{tbl:res_comp} presents the segmentation results on the ACDC and Synapse datasets. Our proposed ASM-UNet achieves the highest average Dice scores on both datasets, outperforming the second-best method by 0.55\% on ACDC and 0.70\% on Synapse. In terms of per-class segmentation performance, ASM-UNet achieves the highest Dice score in the majority of categories across both datasets. Although ASM-UNet does not achieve the top Dice score for MYO (ACDC), LIV (Synapse), and PAN (Synapse), its performance remains within 1\% of the best-performing method, demonstrating its competitive accuracy. A notable exception is the STO category in Synapse, where nnFormer outperforms ASM-UNet by 4.01\%. Nevertheless, ASM-UNet achieves the second-best result in this category and delivers superior performance across most other classes.

\textit{BTMS}: To further validate the effectiveness of ASM-UNet, we conduct additional fine-grained segmentation experiments on our BTMS dataset, which poses significant segmentation challenges. The results are summarized in Table~\ref{tbl:comp_BTMS}, where fine-grained segmentation performance is reported in two forms: ``Avg. Fine (w/o Hard)'' and ``Avg. Fine (w/ Hard)'', with ``Hard'' referring to two particularly challenging categories as defined in Table~\ref{tbl:class_name}. Additionally, we report the coarse-grained performance as ``Avg. Coarse'', which considers only foreground (1) versus background (0) segmentation. 

According to the results, our ASM-UNet achieves the highest performance across all three average metrics: 76.74\% in coarse-grained segmentation, 48.79\% in fine-grained segmentation without hard categories, and 42.57\% with them. Compared to the second-best methods, ASM-UNet achieves improvements in Dice scores of 2.09\% over SegMamba, 1.92\% over CoTr, and 0.78\% over U-Mamba, respectively. Notably, ASM-UNet consistently outperforms all competing methods across most individual anatomical categories, including GB (84.59\%), CBD (48.36\%), CHD (51.60\%), RPHD (32.51\%), and LHD (42.98\%). Here, the Dice score for GB is significantly higher than that of other categories due to its larger volume. This highlights the challenge of segmenting large-scale anatomical structures in FGS task. For the two hard categories, all methods achieve Dice scores below 35\%, with VIT-CUP and Swin-UNet even yielding scores as low as 0\% (marked with ``-'' in Table), primarily due to the tiny volumes and high variability in spatial positions of these structures. 

To investigate the underlying cause of the low accuracy in FGS, we conduct an additional evaluation using CGS, denoted as ``Avg. Coarse''. In this setting, the eight anatomical categories in both the predictions and ground truth are merged into a single foreground class. The results show a substantial increase in Dice scores compared to FGS, indicating that the models are generally capable of detecting the overall biliary system. However, they struggle to accurately distinguish between the individual fine-grained anatomical structures. This highlights the significantly greater challenge posed by FGS relative to common CGS, primarily due to the small size, ambiguous boundaries, and high inter-class similarity among the fine-grained categories.

\begin{figure*}[t]
	\centering
	\includegraphics[width=0.95\textwidth]{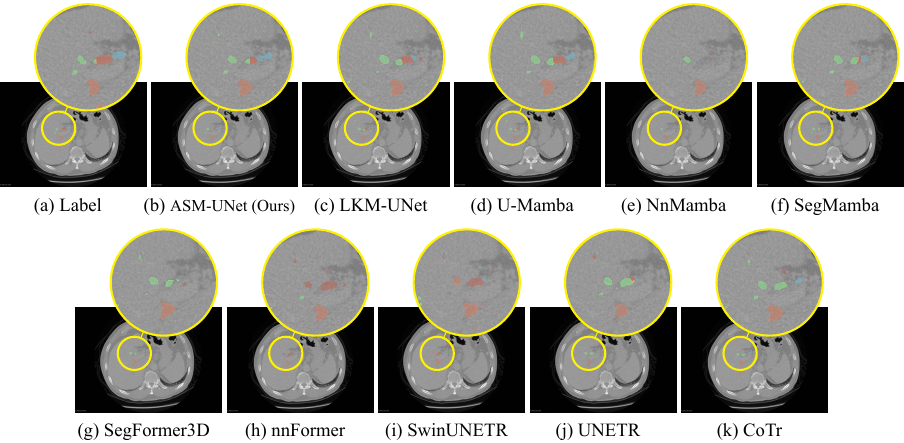}
	\caption{Segmentation results of all compared methods. The region highlighted with a yellow bounding box is enlarged for better visualization.
		\label{fig:slices}}
\end{figure*}

\subsection{Ablation Studies}
\label{subsec:ablation}

Given the challenging nature of the BTMS dataset, we conduct ablation studies to evaluate the contributions of different modules within our model. The results are presented in Table~\ref{tbl:abl_Models}, where ``Mamba'' refers to the baseline model using only the pure Mamba module without scan score enhancements, ``IS'' denotes the incorporation of the individual scan score, and ``GS'' indicates the addition of the group scan score. 
\begin{table}[h]
	\caption{Ablation Studies on Different Modules, Where GS and IS Represent Group and Individual Scan Scores, Respectively; ``Mamba'' Denotes the Fixed Scanning Strategy}
	\footnotesize
	\centering
	\label{tbl:abl_Models} 	
	\renewcommand{\arraystretch}{1.2}
	\setlength{\tabcolsep}{3mm}{
		\begin{tabular}{c|c|c|c|c|c}
			\toprule
			Models& Mamba &IS   &GS   & \makecell{Avg.  Fine \\(w/ Hard)}  &\makecell{Avg.  Fine \\(w/o Hard)}\\
			\midrule
			M1 &  &  &  & 40.23 & 46.68  \\
			M2 & $\checkmark$ &  &  & 40.85 & 46.97 \\
			M3 & $\checkmark$ & $\checkmark$ &  & 41.34 & 47.46 \\
			M4 & $\checkmark$ &  & $\checkmark$ & 41.87 & 48.21 \\
			M5 & $\checkmark$ & $\checkmark$ & $\checkmark$ & \textbf{42.57} & \textbf{48.79}  \\
			\bottomrule
	\end{tabular}}
\end{table}

``M1'' denotes the baseline UNet model, while ``M2'' through ``M5'' correspond to variants equipped with different combinations of the proposed modules. ``M5'', which integrates all components, represents our complete model ASM-UNet. According to the results in Table~\ref{tbl:abl_Models}, we observe a clear performance improvement as each module is incrementally added. Introducing the Mamba module alone (``M2'') improves the Dice score from 40.23\% to 40.85\% on hard categories and from 46.68\% to 46.97\% on non-hard categories, indicating that Mamba provides a solid foundation over the baseline UNet (``M1''). The addition of IS (``M3'') further boosts performance to 41.34\% (hard) and 47.46\% (non-hard), demonstrating its effectiveness by capturing individual differences. When the GS is added (``M4''), the model achieves 41.87\% on hard categories and 48.21\% on non-hard categories, suggesting that utilizing group commonality contributes positively to segmentation quality. Finally, our complete model ASM-UNet (``M5''), which integrates both IS and GS modules, achieves the highest performance with 42.57\% and 48.79\% on hard and non-hard fine-grained segmentation, respectively. These results not only demonstrate the individual effectiveness of each module but also highlight their complementary nature in enhancing the model’s ability to distinguish fine-grained and challenging anatomical structures of biliary system.

\subsection{Parameter Analysis}
\label{subsec:parameter_analysis}
\begin{table}[h]
	\caption{Parameter Analysis on the Impact of Branch Numbers in ASM Block for Segmentation}
	\footnotesize
	\centering
	\label{tbl:abl_branch} 	
	\renewcommand{\arraystretch}{1.2}
	\setlength{\tabcolsep}{2mm}{
		\begin{tabular}{c|c|c}
			\toprule
			Number of Branches & \makecell{Avg.  Fine (w/ Hard)} &\makecell{Avg.  Fine (w/o Hard)}\\
			\midrule
			0& 40.23 & 46.68  \\
			1& 41.91 & 48.70  \\
			2 & 42.01 & 48.46  \\
			3 & \textbf{42.57} & 48.79  \\
			4 & 42.24 & \textbf{49.17} \\
			\bottomrule
		\end{tabular}
	}
\end{table}

As illustrated in Fig.\ref{fig:overview}, multiple branches can be configured to generate adaptive scan scores, which are then fed into the stacked Mamba layers for feature learning. To assess the impact of branch numbers on segmentation performance, we conduct the experiments for parameter analysis, and the results are summarized in Table\ref{tbl:abl_branch}. As shown in the table, introducing multiple branches consistently improves performance over the zero-branch baseline, highlighting the effectiveness of incorporating adaptive scanning representations. In particular, employing three branches yields the highest Dice score (42.57\%) on segmentation with hard categories, while four branches achieve the best result (49.17\%) on non-hard categories. These findings indicate that moderate branching enhances the model's ability to capture diverse semantic patterns. However, increasing the number of branches beyond a certain point may lead to diminishing returns or reduced robustness, particularly in more challenging scenarios. This is because averaging over too many branches may dilute discriminative features, making it harder to preserve the unique characteristics that distinguish hard categories from the others. Therefore, we select three branches as the default setting for fine-grained segmentation experiments on the BTMS dataset.

\subsection{Visual Discussion}

\begin{figure*}[t]
	\centering
	\includegraphics[width=0.985\linewidth]{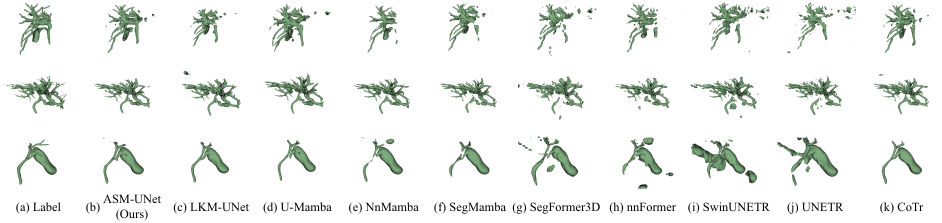}
	\caption{Three examples of coarse-grained segmentation results produced by different methods.
		\label{fig:corase}}
\end{figure*}

\begin{figure*}[t]
	\centering
	\includegraphics[width=0.985\linewidth]{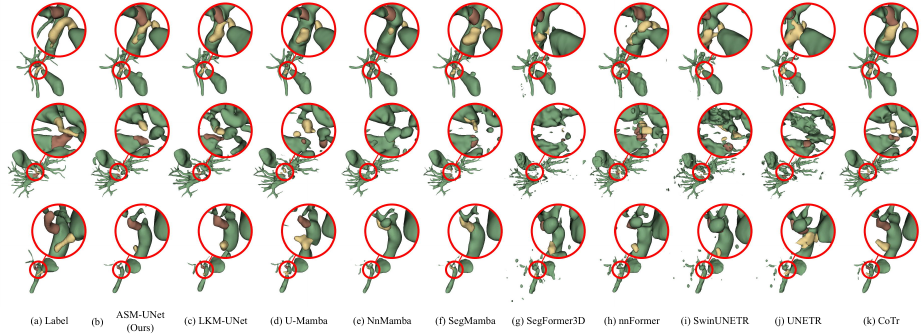}
	\caption{Three examples of segmentation results for the two hard categories produced by different methods.
		\label{fig:hard}}
\end{figure*}

To further analyze the advantage of our method, we show the visual results of FGS and CGS in Fig. \ref{fig:slices} and \ref{fig:corase}, respectively. For FGS, it is evident that our method produces results closely aligned with the ground truth, particularly in delineating the boundaries between RAHD, RHD, and CHD. While LKM-UNet and U-Mamba also achieve promising results, they tend to generate more false positives compared to our ASM-UNet. In contrast, the other methods struggle to distinguish fine-grained categories and often fail to produce reliable segmentation in these regions. For CGS, the reduced task complexity allows all methods to produce visually reasonable results. This further demonstrates that existing methods are generally effective for standard CGS tasks. However, they exhibit limited capability in distinguishing fine-grained anatomical categories. Despite the relative ease of CGS, our method still achieves the best visual performance, delivering continuous and precise segmentation that closely aligns with the ground truth.

Finally, we independently analyze the segmentation results for the two hard categories, CD and RHD, as shown in Fig.\ref{fig:hard}. Despite employing various advanced methods, all models face considerable challenges in correctly identifying and reconstructing these categories. This difficulty stems from their tiny size, irregular morphology, and highly variable spatial distribution. In particular, it remains challenging to distinguish these structures from the background and to preserve their anatomical continuity, such as accurately predicting the CD as a narrow duct connecting the GB and CHD. These findings highlight the limitations of current methods and point to the need for further research on robust, structure-aware segmentation approaches tailored to such fine-grained and spatially inconsistent anatomical components.

\section{Conclusion}
\label{conclusion}

In this paper, we propose a novel Mamba-based model ASM-UNet for FGS tasks. In our approach, a group scan score is introduced to capture group-level commonalities, while an individual scan score is designed to model case-specific variations, which are commonly observed in small-scale anatomical structures. These two scores are combined to generate an adaptive scan score, which guides the Mamba modules in performing feature learning with adaptive scanning, replacing manual and fixed scanning strategies. This design significantly enhances the model's robustness and generalizability in the FGS task. To evaluate the effectiveness of ASM-UNet, we conduct comprehensive comparison studies on two public datasets, ACDC and Synapse. Additionally, to further assess the upper bound of our method's capability, we construct a challenging new dataset, BTMS, which partitions the biliary system into eight fine-grained anatomical structures. Extensive experiments demonstrate that ASM-UNet consistently outperforms state-of-the-art methods in both coarse-grained and fine-grained segmentation tasks. However, all methods, including ours, show limited performance on two particularly challenging categories, CD and RHD, due to their small size and highly variable spatial distribution. Addressing these cases remains an open challenge and an important direction for our future research.

%\appendices

%\bibliographystyle{IEEEtran}
%\bibliography{ref.bib}

\end{document}